\newif\ifblinded
\blindedfalse
\ifblinded
  \documentclass[a4paper,fleqn,doubleblind]{cas-dc}
\else
\documentclass[a4paper,fleqn]{cas-dc}
\fi
\usepackage[utf8]{inputenc}
\usepackage{amssymb}
\usepackage{booktabs}
\usepackage{makecell}
\usepackage{graphicx}
\graphicspath{{figs/}}
\RenewDocumentCommand{\printorcid}{}{}
\usepackage[authoryear]{natbib}

\def\tsc#1{\csdef{#1}{\textsc{\lowercase{#1}}\xspace}}
\tsc{WGM}
\tsc{QE}
\ExplSyntaxOn
\cs_set:Npn \__first_footerline:
  {
    \group_begin:
    \small\sffamily
    \__short_authors: :~
    {\rmfamily\itshape Preprint}
    \group_end:
  }
\ExplSyntaxOff

\begin{document}
\let\WriteBookmarks\relax
% --- Float placement tuning: relax the strict CAS defaults so wide figure*
% floats are placed near their references instead of piling up at the end. ---
\renewcommand{\dbltopfraction}{.9}
\renewcommand{\dblfloatpagefraction}{.8}
\renewcommand{\topfraction}{.9}
\renewcommand{\bottomfraction}{.8}
\renewcommand{\floatpagefraction}{.8}
\renewcommand{\textfraction}{.07}
\setcounter{topnumber}{4}
\setcounter{bottomnumber}{3}
\setcounter{totalnumber}{6}
\setcounter{dbltopnumber}{4}
% Compact but readable separation around both one- and two-column figures.
% Explicit stretch/shrink components let TeX balance the two columns without
% creating conspicuous blocks of white space.
\setlength{\textfloatsep}{10pt plus 2pt minus 2pt}
\setlength{\floatsep}{8pt plus 2pt minus 2pt}
\setlength{\intextsep}{10pt plus 2pt minus 2pt}
\setlength{\dbltextfloatsep}{10pt plus 2pt minus 2pt}
\setlength{\dblfloatsep}{8pt plus 2pt minus 2pt}

\shorttitle{Morphology-Decoupled Segmentation for Tunnel Defects}

\shortauthors{M. Ren et al.}

\title [mode = title]{Morphology-Decoupled Segmentation for Multiple Tunnel Defects Under Complex Environments}  



\ifblinded
\shortauthors{Anonymous authors}
\author{Anonymous authors}
\else

\author[1,5]{Mohan Ren}%[<options>]

% \cormark[1]



\credit{Conceptualization, Methodology, Software, Formal analysis, Validation, Visualization, Writing -- original draft}

\author[1]{Hang Li}%[]
\credit{Methodology, Formal analysis, Validation, Writing -- original draft}

\author[2]{Yifei Liu}%[]
\credit{Conceptualization, Validation, Resources, Writing -- original draft}

\author[3]{Shaowei Wang}%[]
\credit{Methodology, Software, Validation, Visualization, Writing -- original draft}

\author[1]{James O'Kane}
\credit{Conceptualization, Methodology, Validation}

\author[4]{Xuanzhi Wang}%[]
\credit{Formal analysis, Data curation, Writing -- original draft}
\author[4]{Shiyu Zhao}%[]
\credit{Formal analysis, Data curation, Writing -- original draft}

\author[1]{Yumeng Zhang}
\credit{Conceptualization, Supervision, Project administration, Writing -- review \& editing}

\author[4]{Songtao Ji}
\cormark[1]
\ead{songtao.ji@xust.edu.cn}
\credit{Conceptualization, Supervision, Project administration, Funding acquisition, Writing -- review \& editing}

\author[1]{Jurij Karlovsek}%[]
\credit{Conceptualization, Resources, Supervision, Project administration, Funding acquisition, Writing -- review \& editing}

% All affiliations are listed here
\affiliation[1]{organization={School of Civil Engineering, The University of Queensland},
            addressline={St. Lucia}, 
            city={Brisbane},
            postcode={4072}, 
            state={Queensland},
            country={Australia}}

\affiliation[2]{organization={School of Civil and Environmental Engineering, Nanyang Technological University},
            addressline={}, 
            city={Singapore},
            postcode={639798}, 
            country={Singapore}}

\affiliation[3]{organization={School of Civil Engineering, Chang'an University},
            addressline={}, 
            city={Xi'an},
            postcode={710064}, 
            state={Shaanxi},
            country={China}}

\affiliation[4]{organization={Xi'an University of Science and Technology},
            addressline={}, 
            city={Xi'an},
            state={Shaanxi},
            country={China}}

\affiliation[5]{
            organization={Department of Computer Science, University of Exeter},
            addressline={The Queen's Drive},
            city={Exeter},
            postcode={EX4 4QJ},
            state={Devon},
            country={United Kingdom}
}
\fi

\ifblinded\else\cortext[1]{Corresponding author}\fi



\begin{abstract}
Automated tunnel-lining inspection must segment different defect geometries: cracks are thin, elongated, and connectivity-sensitive, whereas leakage, lining falling off, and segment damage are spatially extensive and context-dependent. Existing approaches generally apply a common decoder across defect classes or introduce boundary, topology, and directional mechanisms as isolated enhancements. This paper introduces Morphology-Decoupled Segmentation (MDS), in which defect morphology becomes the organising principle for feature routing, decoding, and interaction. A supervised routing module partitions multiscale encoder features into complementary linear and areal streams. The linear stream employs Deformable Strip Attention with image-conditioned direction hypotheses to preserve crack continuity without assuming fixed horizontal or vertical orientations, while the areal stream combines Pyramid Pooling with feature-pyramid fusion to capture diffuse boundaries and regional context. The representations are coupled through gated cross-morphology attention, allowing complementary information exchange while retaining morphology-specific features. A resolution-adaptive tiling and probability-blending pipeline preserves fine defects across variable-resolution imagery. MDS achieves 81.17\% mIoU on the internal validation set and improves foreground image-level recognition over a backbone-matched SegDINO model, primarily for areal defects. Controlled ablations demonstrate the complementary contributions of routing supervision and cross-stream interaction while revealing a trade-off between areal-class performance and crack specialisation. The submitted model achieved 67.55\% mIoU on an organiser-controlled hidden test, and the team placed second. Selective dynamic INT8 quantisation reduced model storage by $2.83\times$ with negligible accuracy loss.
% Here goes the abstract \nocite{*}%% Remove this line from your manuscript.
\end{abstract}


% Highlights are omitted from this arXiv preview.

%\nocite{*}

% Keywords
% Each keyword is separated by \sep

\maketitle

% Main text
\section{Introduction}\label{sec:introduction}

\subsection{Background}\label{subsec:background}
Rapid urbanisation brings a growing demand for transportation, to which tunnels have proved an effective solution. The mechanical essence of tunnels is intrinsically different from other civil structures, in that tunnels are influenced by the surrounding strata, which pose a large uncertainty. The geological conditions can lead to performance degradation like tunnel defects, including structural deformation \citep{shenLongtermSettlementBehaviour2014} and apparent defects such as cracks, spalling of concrete and water leakage \citep{chiuLiningCrackEvolution2017}. Timely detecting and mapping these defects is an important part of condition assessment \citep{liConditionAssessmentShield2017}.

Manual inspection remains common \citep{insa-iglesias3DVisualInspection2021}, although vehicle-mounted light detection and ranging (LiDAR) systems and cameras now support systematic acquisition \citep{kaartinenLiDARBasedStructuralHealth2022,aiAcquiringSectionalProfile2016}. Some recent developments allow for more efficient and comprehensive data acquisition. Point clouds measure tunnel geometry, while surface photographs support pixel-level mapping of visible defects \citep{zhangInterpretationMethodLaserscanning2024, kaartinenLiDARBasedStructuralHealth2022,aiAcquiringSectionalProfile2016}.

Computer vision methods have been extensively developed across a wide range of fields, shifting from purely handcrafted algorithms to machine-learning-aided methods. Convolutional neural networks (CNNs) and vision transformers have substantially advanced dense prediction \citep{redmonYouOnlyLook2016,liuSwinTransformerHierarchical2021}. The detection of apparent defects from tunnel-surface images is, in this context, essentially a semantic-segmentation task. Visible defects do not necessarily occur in isolation: cracks, leakage, and material loss can coexist within the same inspection image and may overlap or occur in close proximity \citep{wangUAVbasedMultidefectRealtime2025,qinImageSegmentationUsing2024}. A complete condition map must therefore locate and distinguish multiple defect categories simultaneously, rather than process each category through an independent inspection pipeline.

This unified task remains challenging because tunnel imagery contains uneven illumination \citep{huangInspectionEquipmentStudy2017}, background interference, overlapping defect regions, and pronounced geometric differences between thin cracks and spatially extensive damage \citep{wangUAVbasedMultidefectRealtime2025}. These conditions motivate multi-defect segmentation methods that can accommodate distinct defect morphologies while preserving a coherent pixel-level representation of the inspected lining.

\subsection{Related Work}\label{subsec:related-work}
Computer-vision methods have been widely leveraged in tunnel-engineering scenarios, although mostly focusing on specific defect types.

\subsubsection{Single- and Multi-Defect Segmentation for Tunnel Linings}
Crack analysis has progressed from transferred CNN classifiers \citep{brienAutomatedCrackClassification2023} to pixel-level encoder--decoders \citep{houPixellevelCrackSegmentation2024,panHighprecisionSegmentationQuantification2025} and geometry-quantifying acquisition systems \citep{xuAutomaticDetectionTunnel2024}. Recent methods address tunnel curvature, long-range continuity, and tortuous geometry using SLAM, graph or transformer guidance, and dynamic snake convolution \citep{xuTunnelCrackAssessment2025,liuCGVNetTunnelLining2025,duanDSYOLODynamicSnake2025}. These developments explicitly exploit crack geometry and improve continuity, but remain centred on one defect type. Their inductive biases do not directly address diffuse leakage boundaries or block-like material loss in a shared label space.

Leakage varies with illumination, wetness, staining, and diffuse boundaries. Existing methods combine tiled large-image inference, explainable U-Net variants, super-resolution attention, or low-light adaptation \citep{tanLightweightMetroTunnel2023,fengDeepLearningVisual2023,chenIntelligentIdentificationTunnel2024,wangImageSegmentationTunnel2024}. The field is consequently moving from binary extraction towards robust and fine-grained parsing. Spalling has received less attention despite its connection to durability degradation; LiDAR unrolling with U-Net segmentation has been used to estimate masonry-spalling severity and depth \citep{smithAutomatedMasonrySpalling2024}, and reviews identify it as an emerging multi-defect target \citep{zhouIntelligentRecognitionTunnel2025,wangDeepLearningBased2026}.

These pipelines are effective within their target regimes, but direct integration is difficult: crack models favour thin connected responses, whereas leakage models favour spatial completeness and may smooth local variation. Applying either bias uniformly can disadvantage the other morphology.

Unified approaches remain comparatively scarce. MC-TLD combines multiscale attention and contextual enhancement for environmental interference and multi-scale targets \citep{zhouAutomaticSegmentationTunnel2023}, while a two-stage framework separates defective-image screening from downstream recognition of leakage, cracks, and spalling \citep{fengTwostepDeepLearningbased2024}. This separation reflects the difficulty of resolving diverse appearances in one end-to-end pathway. UAV-based detection further exposes co-existing defects, uneven lighting, small targets, and spalling near bolts \citep{wangUAVbasedMultidefectRealtime2025}. SAM-based instance segmentation and ultra-high-resolution panoptic modelling broaden category coverage \citep{yeSambasedInstanceSegmentation2024,yeMaintenanceorientedTunnelDigital2026}. Nevertheless, these formulations do not explicitly allocate representation capacity according to linear and areal morphology, and reliable segmentation remains difficult when such defects overlap \citep{qinImageSegmentationUsing2024}.

\subsubsection{Morphology-Aware and Decoupled Segmentation}
Outside tunnel-specific research, the computer-vision literature has explored several strategies for incorporating shape or topology awareness into semantic segmentation. DecoupleSegNet separates body and edge features by flow-field warping, and Gated-SCNN adds a gated boundary stream \citep{liDecoupleSegNet2020,takikawaGatedSCNNGatedShape2019}. Dynamic Routing selects scale-dependent computational paths at image level \citep{liLearningDynamicRouting2020}. clDice and Betti Matching instead impose topological constraints through the loss \citep{shitClDiceNovelTopologyPreserving2021,stuckiTopologicallyFaithfulImage2022}. Strip Pooling and Axial Attention provide directional context along fixed axes \citep{houStripPoolingRethinking2020,wangAxialDeepLabStandAloneAxialAttention2020}.

Together, these studies show how geometric characteristics can inform both feature processing and supervision. Boundary streams help delineate the contours of spatially extensive regions, topology-aware losses preserve the connectivity of elongated structures, directional operators aggregate evidence along anisotropic patterns, and dynamic routing separates features that require different receptive fields. However, these mechanisms are generally applied in isolation or uniformly across semantic categories. They do not organise a unified tunnel-defect model around the coexistence of linear cracks and areal damage, in which each morphology receives dedicated feature routing and decoding before controlled cross-morphology interaction.

Despite this progress, the marked geometric disparity between linear cracks and areal damage is rarely used to organise feature routing and decoding within a unified tunnel-inspection model. The problem is especially consequential when both regimes coexist in one image and must be segmented without sacrificing crack continuity or regional context.

\subsection{Research Overview}\label{subsec:research-overview}
Motivated by this gap, the present research analyses 1{,}100 tunnel-lining images and 3{,}089 annotated defect instances in terms of category distribution, geometry, and native resolution. It then studies the separate processing of linear cracks and areal defects through Morphology-Decoupled Segmentation (MDS). Unlike approaches that append a morphology-aware loss or attention operator to an otherwise shared segmentation pathway, MDS uses morphology to determine how encoder features are routed, how they are decoded by specialised branches, and how the resulting representations interact before prediction. The framework combines a Morphological Routing Module (MRM), a Deformable Strip Attention (DSA) linear decoder, a Pyramid Pooling Module--Feature Pyramid Network (PPM-FPN) areal decoder, and a Cross-Morphology Interaction Module (CMIM). The study further specifies a resolution-adaptive data pipeline, morphology-aware supervision, and selective dynamic 8-bit integer (INT8) conversion. Evaluation covers representative baselines, controlled ablations, morphology-specific diagnostics, qualitative cases, and an organiser-controlled hidden test.

In the 2026 Intelligent Multi-defect Detection for Tunnel Linings in Complex Environments competition, the submitted model achieved 67.55\% seven-class mIoU on the organiser-controlled hidden test, and the team placed second.

The remainder of the paper follows this research sequence. Section~\ref{sec:data-pipeline} characterises the dataset and presents the data pipeline. Section~\ref{sec:method} describes the MDS architecture and training objective. Section~\ref{sec:experimental-setup} defines the evaluation and compression protocols, and Section~\ref{sec:results} reports the experimental results. Section~\ref{sec:discussion} discusses the observed behaviour and limitations, followed by the conclusion in Section~\ref{sec:conclusion}.

\section{Dataset Characteristics and Morphology-Guided Data Pipeline}\label{sec:data-pipeline}

This section examines the dataset as the empirical basis for the proposed methodology and data pipeline. The imagery captures several challenges commonly reported in practical tunnel-lining inspection, including uneven illumination, surface contamination, background clutter, substantial resolution variation, class imbalance, and the coexistence of morphologically distinct defects \citep{huangInspectionEquipmentStudy2017,wangUAVbasedMultidefectRealtime2025}. We therefore focus on four observations that have direct modelling consequences: the semantic composition and multi-defect content of the dataset, the instance- and pixel-level imbalance that shapes optimisation, the morphological separation between linear and areal defects that motivates specialised decoding, and the resolution variation that requires adaptive tiling.

\subsection{Dataset Overview}\label{subsec:dataset-overview}
The dataset contains 1,100 PNG tunnel-lining images with LabelMe polygon annotations. The six foreground classes are crack, black-stain leakage (Leakage~B), white efflorescence leakage (Leakage~W), algae-associated leakage (Leakage~G), lining falling off, and segment damage. With the unlabelled tunnel surface treated as background, the task has seven semantic classes.

\begin{figure*}[pos=t]
  \centering
  \includegraphics[width=\linewidth]{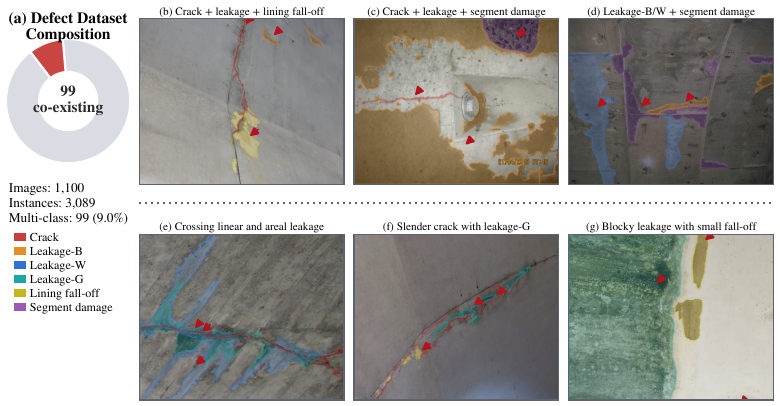}
  \caption{Dataset composition and multi-defect examples. (a) Of 1{,}100 images and 3{,}089 instances, 99 images (9.0\%) contain multiple classes. (b) Crack, leakage, and lining fall-off. (c) Crack, leakage, and segment damage. (d) Leakage~B/W with segment damage. (e) Crack crossing areal leakage. (f) Crack with Leakage~G. (g) Leakage~G with small fall-off regions. Colours follow the legend; arrowheads mark selected defects. Images provided by the organiser.}
  \label{fig:coexistence}
\end{figure*}

The images contain uneven illumination, background clutter, surface contamination, carbonation texture, and occlusion. Beyond isolated defects, 99 images (9.0\%) contain two or more co-existing defect categories (Figure~\ref{fig:coexistence}). The observed combinations include cracks with lining fall-off, Leakage~B with segment damage, and other multi-defect configurations. Producing a complete condition map for these scenes requires one model to distinguish defects with different spatial extents and boundary characteristics. The multi-defect content therefore motivates a unified segmentation framework, while the geometric differences among its categories motivate the morphology-decoupled design developed in Section~\ref{sec:method}.

\subsection{Statistical Class Imbalance}\label{subsec:class-imbalance}

\begin{table}[!ht]
\centering
\caption{Class distribution of tunnel lining defects.}
\label{tab:class_distribution}
\renewcommand{\arraystretch}{1.15}
\setlength{\tabcolsep}{11pt}
\begin{tabular}{cccc}
\toprule
Class & Instances & \makecell{Instance\\Share} & \makecell{Pixel\\Share} \\
\midrule
Leakage B & 1,412 & 45.71\% & 72.21\% \\
Crack & 759 & 24.58\% & 0.36\% \\
\makecell{Lining falling\\off} & 556 & 18.00\% & 4.73\% \\
\makecell{Segment\\damage} & 213 & 6.90\% & 16.59\% \\
Leakage W & 91 & 2.95\% & 1.88\% \\
Leakage G & 58 & 1.88\% & 4.23\% \\
\bottomrule
\end{tabular}
\end{table}

A statistical analysis of 3,089 annotated defect instances reveals severe imbalance at both the instance and pixel levels. Table~\ref{tab:class_distribution} summarises the most relevant distributional evidence. At the instance level, Leakage~B is the dominant annotated category, whereas Leakage~G is the rarest. At the pixel level, however, the most critical observation is the divergence between crack frequency and crack area: cracks are the second most frequent defect by instance count but contribute only 0.36\% of annotated pixels.

This distribution exposes two different imbalance mechanisms. Instance-level imbalance creates a long-tailed classification problem among defect categories, while pixel-level imbalance directly affects segmentation optimisation. The pixel-level imbalance ratio reaches 201.57 between Leakage~B and crack, meaning that ordinary per-pixel losses can be dominated by large leakage regions while under-penalising missed crack pixels. The instance and pixel shares in Table~\ref{tab:class_distribution} make this reversal explicit: cracks rank second by instance count but last by pixel area. This motivates the overlap-aware and hard-example-focused training objective formalised in Section~\ref{subsec:training-objective}. The instance--pixel mismatch is therefore not only a data-imbalance issue but also an optimisation issue: rare, thin, and boundary-ambiguous defects contribute weak gradient signals unless the training objective explicitly compensates for them.

\subsection{Morphological Heterogeneity}\label{subsec:morphology}
\begin{figure}[pos=t]
  \centering
  \includegraphics[width=\linewidth]{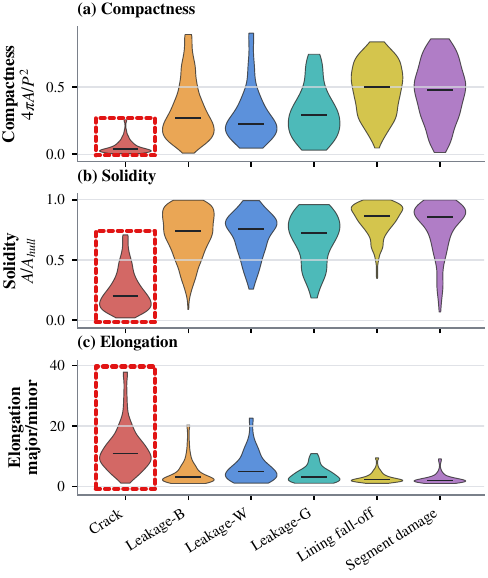}
  \caption{Descriptor distributions by class: (a) compactness, (b) solidity, and (c) elongation. Violin width shows density, black bars mark medians, and red frames highlight cracks, which separate from the five areal categories.}
  \label{fig:violin}
\end{figure}

Beyond frequency imbalance, the dataset exhibits a more fundamental morphological heterogeneity. The dataset contains 3,089 annotated defect polygons, with each LabelMe polygon treated as one instance for dataset-level statistics. Degenerate polygons with fewer than three points, area below 5 pixels, or non-positive perimeter are excluded from numerical descriptor computation. Compactness is $4\pi A/P^2$, solidity is the polygon area divided by its convex-hull area, and elongation is the square root of the ratio of the major and minor eigenvalues of the second central-moment matrix. The class-wise distributions of compactness, solidity, and elongation in Figure~\ref{fig:violin} make the separation between thin cracks and the five areal categories directly visible.

\paragraph{Areal defect regime.} Across the five areal categories, median solidity ranges from 0.723 to 0.866, median compactness from 0.226 to 0.499, and median elongation from 2.03 to 4.95. Their segmentation is complicated by diffuse boundaries and confusion with staining, shadows, and concrete weathering. These observations motivate the context-aggregating areal decoder of Section~\ref{subsec:areal-decoder}.

\paragraph{Linear defect regime.} Cracks occupy a sharply different region of shape-descriptor space, with median compactness nearly an order of magnitude lower than the areal classes ($0.036$ vs.\ $0.226$--$0.499$) and median elongation of $10.88\times$. Their thin, fragmented geometry makes them highly sensitive to small lateral prediction errors. For a crack of width $w$ and lateral displacement $\delta$, a simplified rectangular model gives the intersection over union (IoU) as
\begin{equation}
\mathrm{IoU}(w,\delta) = \max\!\left(0,\ \frac{w-\delta}{w+\delta}\right).
\label{eq:crack_iou}
\end{equation}
As shown in Table~\ref{tab:crack_sensitivity}, even a $1$-pixel shift can halve the IoU of a $3$-pixel-wide crack. This motivates both the morphology-specific linear decoder and the direction-adaptive strip attention mechanism of Section~\ref{subsec:dsa-decoder}.

\begin{table}[!ht]
\centering
\caption{Crack IoU sensitivity to lateral spatial shifts, computed from Eq.~\eqref{eq:crack_iou}.}
\label{tab:crack_sensitivity}
\renewcommand{\arraystretch}{1.15}
\setlength{\tabcolsep}{5.5pt}
\begin{tabular}{ccccl}
\toprule
\makecell{Crack\\Width} & \makecell{IoU\\0\,px} & \makecell{IoU\\1\,px} & \makecell{IoU\\2\,px} & Interpretation \\
\midrule
2\,px  & 1.000 & 0.333 & 0.000 & Almost no tolerance \\
3\,px  & 1.000 & 0.500 & 0.200 & 1\,px error halves IoU \\
5\,px  & 1.000 & 0.667 & 0.429 & Still highly sensitive \\
10\,px & 1.000 & 0.818 & 0.667 & Moderate tolerance only \\
\bottomrule
\end{tabular}
\end{table}

\subsection{Resolution Variations}\label{subsec:resolution-variation}
The official images exhibit large resolution variation, creating a trade-off between preserving fine defect details and maintaining computational feasibility. A fixed global resize to $512\times512$ would be particularly harmful for cracks: thin structures may collapse to sub-pixel width, while aspect-ratio distortion can alter their elongated morphology. At the same time, directly processing every raw high-resolution image is impractical under the model-size and deployment constraints considered in this work. Images are therefore handled using resolution-adaptive groups rather than a single forced-resize rule; Table~\ref{tab:resolution_groups} summarises the adopted grouping logic and its modelling motivation.

\begin{table}[!ht]
\centering
\caption{Resolution groups and their processing strategies.}
\label{tab:resolution_groups}
\renewcommand{\arraystretch}{1.2}
\setlength{\tabcolsep}{5pt}
\begin{tabular}{@{}p{1.3cm}p{1.4cm}p{2.2cm}p{1.6cm}@{}}
\toprule
Group & Pixel Range & Processing & Stride \\
\midrule
Tiny & $<$200\,K & Resize to $512^2$ & --- \\
Mobile & 200\,K--2\,M & $512^2$ tiles & 384 \\
DSLR & 2\,M--8\,M & $640^2$ tiles $\to$ input & 400 \\
HighRes & $>$8\,M & $768^2$ tiles $\to$ input & 384 \\
\bottomrule
\end{tabular}
\end{table}

\subsection{Resolution-Adaptive Input and Dataset Construction}\label{subsec:input-pipeline}
\begin{figure*}[pos=t]
  \centering
  \includegraphics[width=\linewidth]{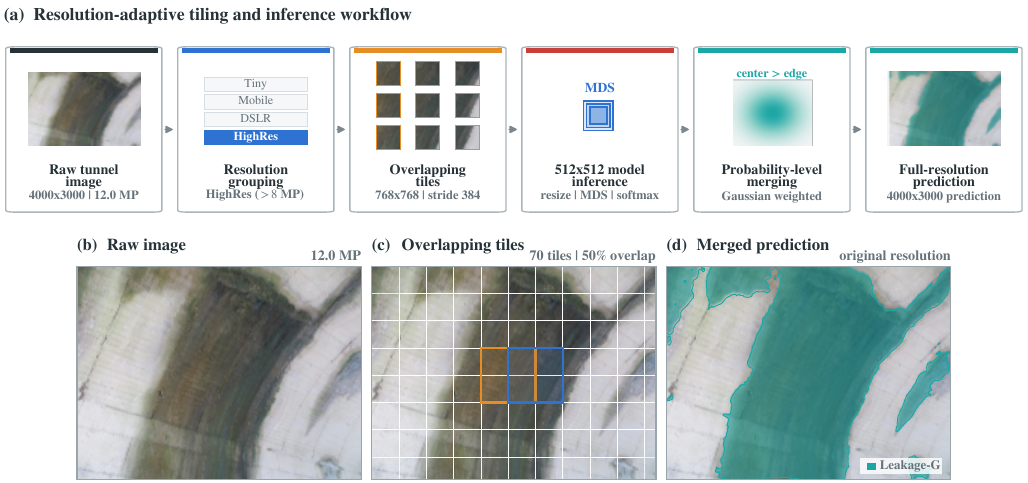}
  \caption{Resolution-adaptive inference. (a) Workflow from resolution grouping and overlapping tiles to $512\times512$ inference, probability merging, and reconstruction; HighRes uses $768\times768$ tiles with stride 384. (b) A 12.0-MP image. (c) Its 70 tiles at 50\% overlap. (d) Gaussian-weighted full-resolution Leakage~G prediction.}
  \label{fig:tiling}
\end{figure*}

To prevent information leakage between training and validation, the 1{,}100 source images are split at the image level \emph{before} patch generation using an 80:20 ratio, yielding 880 training images and 220 validation images. Raw images are then grouped by native resolution and processed using the tile sizes and strides in Table~\ref{tab:resolution_groups}. Tiny images are resized directly; larger images are reflect-padded at the right and bottom edges and decomposed into overlapping tiles. Images are resized with bilinear interpolation and masks with nearest-neighbour interpolation to a $512\times512$ network input. The resulting patch-level dataset contains 14{,}739 training patches and 3{,}749 validation patches, with the validation patches retaining their source-image organisation for full-resolution evaluation.

For tiled images, the same crop is applied to the RGB image and class-index mask, and all background-only patches are retained during conversion. During both MDS and SegDINO training, a replacement-weighted sampler assigns weight 3.0 to samples containing cracks and 2.0 to samples containing Leakage~W. Applying the same sampling protocol prevents rare-class exposure from differing systematically between the principal baseline and MDS.

\subsection{Data Augmentation}\label{subsec:data-augmentation}
Training augmentation comprises random resized cropping (scale 0.4--1.0; aspect ratio 0.75--1.33), horizontal and vertical flips, $90^\circ$ rotations, elastic deformation, Gaussian noise, and a randomly selected photometric transformation (brightness--contrast, gamma, or hue--saturation--value). These transformations increase geometric and photometric diversity and improve robustness to the illumination variability encountered in tunnel imagery. Mosaic and shear augmentation are not used.

\subsection{Full-Resolution Inference}\label{subsec:tiled-inference}
During raw-resolution inference, overlapping tile predictions are projected back to the original image coordinate system and merged at the probability level. Let $P_{k,c}(x)$ denote the predicted probability of class $c$ at pixel $x$ from tile $k$, and let $w_k(x)$ denote the corresponding Gaussian tile weight, which decays toward the tile borders. The merged probability and the final label are obtained by
\begin{align}
P_c(x) &= \frac{\sum_k w_k(x)\,P_{k,c}(x)}{\sum_k w_k(x) + \epsilon},\label{eq:tile_blend}\\
\hat{y}(x) &= \arg\max_c P_c(x).\label{eq:tile_label}
\end{align}
Here each tile uses a separable Gaussian window with standard deviation one quarter of the tile size, and $\epsilon=10^{-8}$. Predictions are averaged with those obtained from a horizontal flip before blending. Equation~\eqref{eq:tile_blend} reduces tile-boundary discontinuities while retaining the raw-image coordinate system used for evaluation.

\section{Morphology-Decoupled Segmentation}\label{sec:method}

\subsection{Overall Architecture}\label{subsec:architecture}
MDS is designed around the morphological separation established in Section~\ref{sec:data-pipeline}: thin linear cracks require continuity-preserving modelling, whereas leakage, lining falling off, and segment damage require region-level contextual aggregation. The framework therefore decouples feature decoding into a linear stream and an areal stream while preserving controlled interaction between them.

\begin{figure*}[pos=t]
  \centering
  \includegraphics[width=\linewidth]{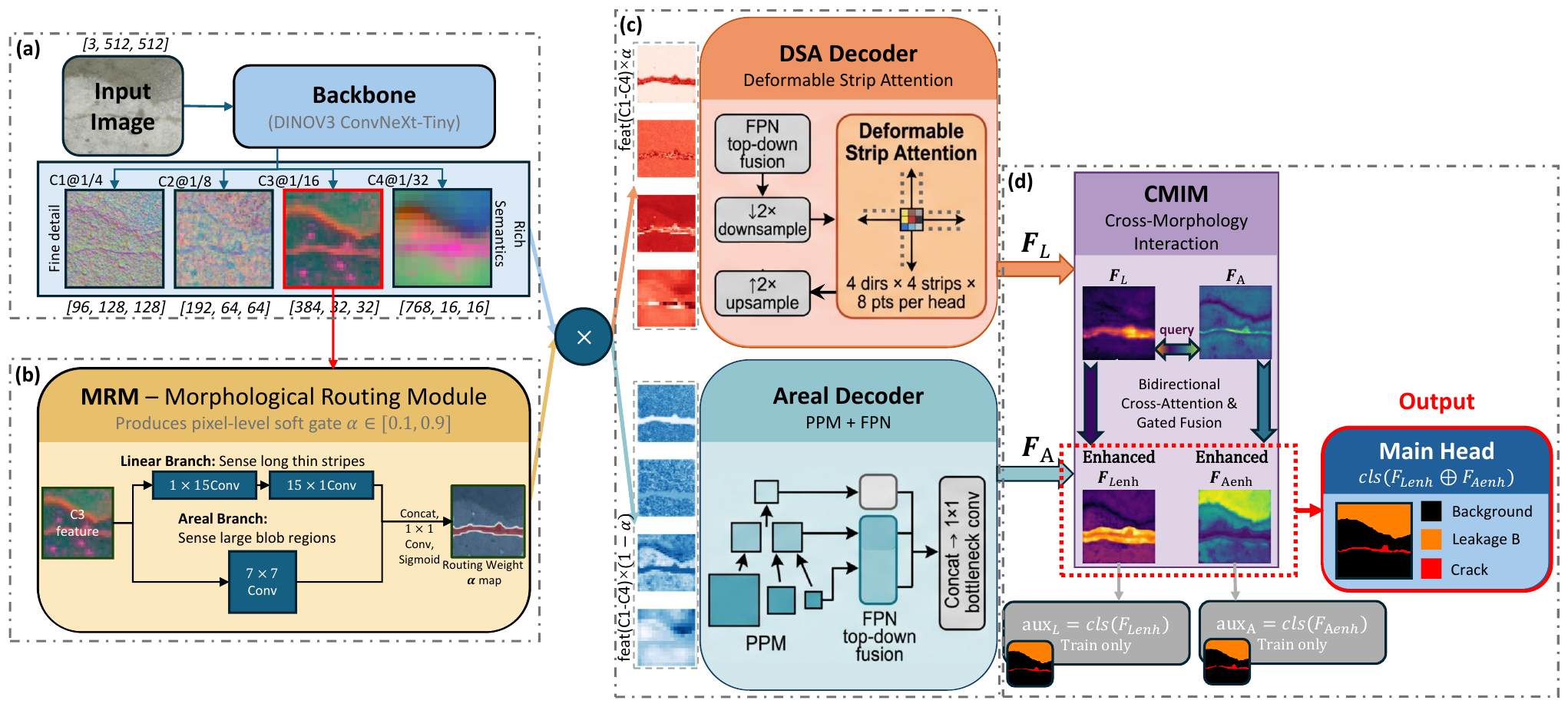}
  \caption{MDS architecture. (a) DINOv3-distilled ConvNeXt-Tiny backbone and features C1--C4. (b) MRM routing into complementary linear and areal streams. (c) DSA and PPM-FPN decoders producing $F_L$ and $F_A$. (d) CMIM cross-attention, gated refinement, main prediction, and training-only auxiliary heads.}
  \label{fig:arch}
\end{figure*}

Figure~\ref{fig:arch} organises the information flow into four stages. First, the DINOv3-distilled ConvNeXt-Tiny backbone extracts a shared multiscale feature pyramid at strides 4, 8, 16, and 32 (Figure~\ref{fig:arch}(a)). Second, the MRM predicts a soft routing map $\alpha\in[0.1,0.9]$ and uses it to partition every scale into complementary linear and areal streams (Figure~\ref{fig:arch}(b)). Third, the DSA and PPM-FPN decoders process the two streams according to their distinct morphological requirements (Figure~\ref{fig:arch}(c)). Finally, CMIM permits gated information exchange without replacing either specialised representation, after which the enhanced features are fused for the final $7$-class prediction (six defect categories plus background; Figure~\ref{fig:arch}(d)).

\subsection{Backbone: DINOv3-Distilled ConvNeXt-Tiny}\label{subsec:backbone}
The backbone stage in Figure~\ref{fig:arch}(a) converts the input image into a shared hierarchy from which both morphology-specific streams are derived. Given an input image, it produces four feature maps $F^{(1)}$--$F^{(4)}$, denoted C1--C4, at strides $4$, $8$, $16$, and $32$ with channel dimensions $96$, $192$, $384$, and $768$, respectively. C1 and C2 retain fine boundary and texture cues needed to localise narrow cracks, whereas C3 and C4 provide progressively broader semantic context for spatially extensive defects. All four maps are passed to the routing stage in Figure~\ref{fig:arch}(b), where their morphology-specific allocation is determined.

We instantiate this stage with the DINOv3-distilled ConvNeXt-Tiny backbone \citep{simeoniDINOv32025}; related DINOv3 features transfer effectively to surface-defect segmentation \citep{xueDINov3UNetPretrainedVision2025}. This backbone combines self-supervised dense representations with a convolutional hierarchy, and its four stages naturally support FPN-style decoding. The linear layers within its multilayer perceptron (MLP) blocks are also eligible for the selective dynamic INT8 quantisation described in Section~\ref{subsec:quantisation-protocol}. SegDINO, the principal baseline, uses the same DINOv3 ConvNeXt-Tiny encoder and rare-class sampling protocol. The controlled comparison and training details are defined in Section~\ref{sec:experimental-setup}.

\subsection{Morphological Routing Module}\label{subsec:mrm}
The MRM addresses incompatible geometric cues entangled in the shared pyramid. C3 supplies the semantic representation from which one spatial routing map is predicted; its resized forms weight C1--C4, producing complementary soft pyramids $F_L^{(k)}$ and $F_A^{(k)}$ for the DSA and PPM-FPN decoders. Unlike a hard assignment, this construction allows both branches to retain evidence where morphology is uncertain.

\begin{figure}[pos=t]
  \centering
  \includegraphics[width=\linewidth]{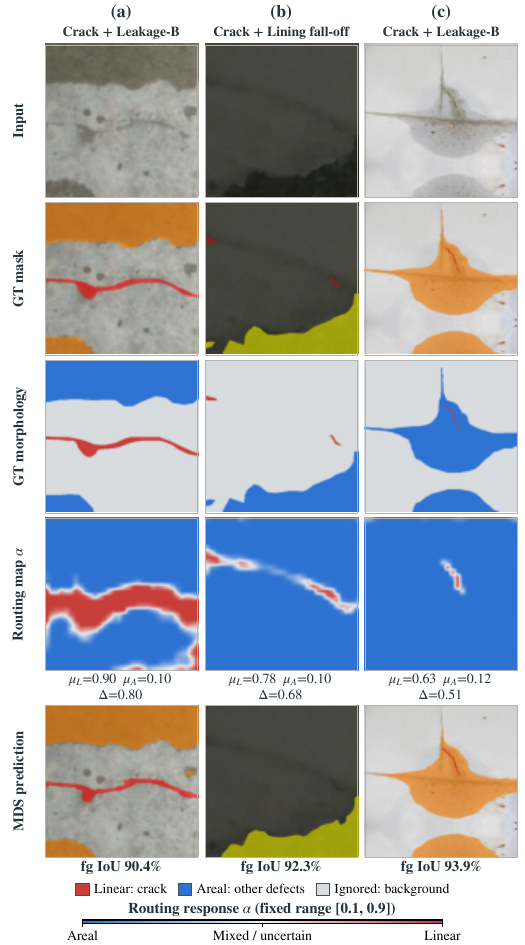}
  \caption{Supervised routing for (a,c) crack with Leakage~B and (b) crack with lining fall-off. Rows show input, ground truth, morphology target, routing map $\alpha$, and MDS prediction. Red/blue denote linear/areal targets; grey background is ignored. Low/high $\alpha$ favours the areal/linear stream. $\mu_L$, $\mu_A$, $\Delta$, and foreground IoU are reported below.}
  \label{fig:routing}
\end{figure}

Concretely, the router takes the C3 feature $F^{(3)}\in\mathbb{R}^{384\times H/16\times W/16}$. Its linear branch applies $1\times15$ and $15\times1$ convolutions, whereas its areal branch uses a $7\times7$ convolution. Both branches produce 64 channels; their concatenation is projected to one logit channel. The sigmoid output is clipped to prevent either stream from being completely suppressed:
\begin{equation}
\alpha = \operatorname{clip}\!\left(\sigma\!\left(f_{\text{MRM}}(F^{(3)})\right),0.1,0.9\right).
\label{eq:mrm_gate}
\end{equation}
One map is shared across the feature pyramid. With $\alpha^{(k)}$ denoting its bilinear resize to scale $k$, the routed features are
\begin{equation}
F_{L}^{(k)} = \alpha^{(k)} \odot F^{(k)}, \qquad F_{A}^{(k)} = (1-\alpha^{(k)})\odot F^{(k)}.
\label{eq:mrm_split}
\end{equation}
Element-wise multiplication is denoted by $\odot$. The streams are complementary but remain soft because of the clipping interval. Routing applies to every location; only its loss excludes background and ignored pixels. Figure~\ref{fig:routing} illustrates the morphology partition that supervises the MRM on co-existing cases: crack pixels form the linear target, areal-defect pixels form the areal target, and background pixels are ignored. The predicted routing weights remain soft, as defined in Eq.~\eqref{eq:mrm_split}.

\subsection{Linear DSA Decoder}\label{subsec:dsa-decoder}
The linear decoder in Figure~\ref{fig:arch}(c) addresses the localisation and connectivity sensitivity of narrow cracks. As shown in Section~\ref{subsec:morphology}, a lateral prediction shift of just $1$--$2$ pixels can halve or eliminate the IoU of a narrow crack. Standard convolutional decoders aggregate features within local isotropic neighbourhoods (e.g.\ $3\times3$ kernels), which may capture mostly background when centred on a thin crack and do not explicitly follow the directional continuity that defines crack morphology. Faithful crack segmentation therefore requires not only detecting foreground pixels but tracing their continuous path across the lining surface, since fragmented predictions risk underestimating crack severity and misleading maintenance decisions.

The decoder receives the routed linear maps $F_L^{(1)}$ through $F_L^{(4)}$. An FPN first fuses them into a $32$-channel feature map at $H/4$ resolution, which is average-pooled to $H/8$ for Deformable Strip Attention. Four attention heads each predict four unit direction hypotheses from global image-level feature statistics and aggregate evidence along oriented strips rather than fixed horizontal or vertical axes. The resulting linear representation is returned to $H/4$ resolution as $F_L$ and passed to CMIM in Figure~\ref{fig:arch}(d). For every query position and direction $\mathbf d_{h,m}$, eight points are sampled in the normalised feature grid:
\begin{equation}
\begin{aligned}
\mathbf{p}_{h,m,j}&=\operatorname{clip}\!\left(\mathbf p+0.5t_j\mathbf d_{h,m},-1,1\right),\\
t_j&\in\operatorname{linspace}(-0.5,0.5,8).
\end{aligned}
\label{eq:dsa_sampling}
\end{equation}
Sampling uses bilinear interpolation with border padding. Attention is normalised over the 32 sampled keys per head, followed by group normalisation and a residual connection. The result is upsampled to $H/4$ before CMIM.

From a structural-inspection perspective, this mechanism can be understood as a learned analogue of the directional scanning that engineers perform when tracing a crack by eye: first estimate the crack's dominant orientation, then concentrate analysis along that direction. It differs from fixed-axis pooling methods, which assume horizontal or vertical alignment, and from isotropic deformable attention, which scatters sampling points without directional structure. The per-image global direction prediction reflects the empirical observation that tunnel cracks, while varying in orientation across images, tend to exhibit locally consistent directionality within a single inspection frame.

\subsection{Areal PPM-FPN Decoder}\label{subsec:areal-decoder}
The areal decoder in Figure~\ref{fig:arch}(c) addresses defects whose evidence is distributed across regions rather than concentrated along a direction. It receives the routed areal pyramid $\{F_A^{(k)}\}_{k=1}^{4}$, in which the deeper maps provide category-level context and the shallower maps retain boundary and texture detail. Pyramid Pooling first aggregates the deepest routed feature $F_A^{(4)}$ at $1\times1$, $2\times2$, $4\times4$, and $8\times8$ scales so that diffuse or spatially extensive regions can be interpreted using multiple contextual extents. Each pooled map is projected to 32 channels and upsampled. An FPN then fuses the context-enriched $F_A^{(4)}$ with the other routed areal scales, producing $F_A\in\mathbb{R}^{32\times H/4\times W/4}$ for CMIM in Figure~\ref{fig:arch}(d). Table~\ref{tab:decoder_asymmetry} summarises the resulting decoder asymmetry.

\begin{table*}[pos=t]
\centering
\caption{Deliberate design asymmetry between the two decoders, reflecting the morphological dichotomy established in Section~\ref{sec:data-pipeline}.}
\label{tab:decoder_asymmetry}
\renewcommand{\arraystretch}{1.2}
\begin{tabular}{lll}
\toprule
 & DSA Decoder (Linear) & Areal Decoder \\
\midrule
Target morphology & Cracks ($>10\times$ elong.) & Leakage, spalling, damage \\
Core mechanism & Strip attention & Pyramid Pooling \\
Receptive field & Directional, adaptive & Isotropic, multi-scale \\
Output & $[B,32,\tfrac{H}{4},\tfrac{W}{4}]$ & $[B,32,\tfrac{H}{4},\tfrac{W}{4}]$ \\
\bottomrule
\end{tabular}
\end{table*}

\subsection{Cross-Morphology Interaction Module}\label{subsec:cmim}
In real tunnel environments, cracks and areal defects are not always independent phenomena. Water ingress commonly follows pre-existing structural discontinuities, and material deterioration may radiate outward from crack networks. Processing the two morphologies in complete isolation would forfeit relational evidence in such co-occurrence regions, whereas unconstrained fusion could erase the specialisation established by routing and branch-specific decoding. CMIM in Figure~\ref{fig:arch}(d) therefore exchanges context only after each decoder has formed its own representation and controls how much of that context each stream absorbs.

The module takes the decoded $32$-channel maps $F_L$ and $F_A$ at $H/4$ resolution. It average-pools both maps by a factor of four, performs bidirectional cross-attention between the resulting $H/16$ representations, and upsamples the two cross-stream contexts to the decoder resolution. Independent sigmoid gates then determine, at each spatial position and channel, how much context is added to the corresponding original feature through a residual path. The outputs $\widetilde{F}_L$ and $\widetilde{F}_A$ consequently retain their branch-specific identity while incorporating only gated complementary evidence, and they are forwarded to the segmentation heads. Two independent instances of a multi-head attention operator instantiated with one head (MHA; embedding width 32 and head dimension 32, with no shared parameters) produce the pooled cross-stream contexts:
\begin{align}
\bar{C}_{L} &= \operatorname{MHA}(Q=\bar{F}_{L},\,K=\bar{F}_{A},\,V=\bar{F}_{A}),\label{eq:cmim_context_l}\\
\bar{C}_{A} &= \operatorname{MHA}(Q=\bar{F}_{A},\,K=\bar{F}_{L},\,V=\bar{F}_{L}).\label{eq:cmim_context_a}
\end{align}
The pooled contexts are bilinearly upsampled to the decoder resolution, $C_L=\operatorname{Up}_4(\bar{C}_L)$ and $C_A=\operatorname{Up}_4(\bar{C}_A)$. Equations~\eqref{eq:cmim_context_l} and~\eqref{eq:cmim_context_a} define the two directions of information exchange. Rather than directly injecting this context, a learned gate controls the degree of absorption at each spatial position and channel. For the linear stream, a sigmoid gate is computed from the concatenation of the original features and the upsampled cross-attention context, and the enhanced feature is formed as an additive, gated residual followed by layer normalisation (LN):
\begin{align}
g_{L} &= \sigma\!\left(W_{L}\,[\,F_{L};\,C_{L}\,]\right),\label{eq:cmim_gate_l}\\
\widetilde{F}_{L} &= \operatorname{LN}\!\left(F_{L} + g_{L}\odot C_{L}\right).\label{eq:cmim_fusion_l}
\end{align}
Equation~\eqref{eq:cmim_gate_l} determines how much cross-stream context is admitted, and Eq.~\eqref{eq:cmim_fusion_l} applies it as a normalised residual. The areal path uses the symmetric operation with $L$ and $A$ exchanged and an independently parameterised $64\!\to\!32$ linear-sigmoid gate $W_A$; channel-wise LN follows both residual additions. Thus, dense cross-attention is confined to $1/16$ resolution, while gated refinement is performed at $1/4$ resolution before the two specialised outputs are passed to the classification heads.

\subsection{Main and Auxiliary Segmentation Heads}\label{subsec:classification-heads}
As depicted in Figure~\ref{fig:arch}(d), the model contains one main segmentation head and two training-only auxiliary segmentation heads. Following CMIM, the main head concatenates the two 32-channel feature maps at $H/4$ resolution, and a $1\times1$ convolution maps the resulting 64 channels to seven logits. The linear and areal auxiliary heads independently project $\widetilde{F}_{L}$ and $\widetilde{F}_{A}$ from 32 channels to the same seven logits. None of the heads contains an intermediate non-linearity.

Both auxiliary heads receive the complete seven-class target and provide branch-specific deep supervision. Morphological differentiation is governed by feature routing and routing supervision, while the auxiliary losses strengthen the individual decoder representations during training. The auxiliary heads are discarded at inference. The logits from the main head are bilinearly upsampled to the input resolution before the channel-wise $\arg\max$.

\subsection{Training Objective}\label{subsec:training-objective}
Let $N$ denote the number of valid pixels, $C=7$, $p_{i,c}$ the softmax probability, and $g_{i,c}$ the one-hot target. Pixels labelled 255 are ignored. In the order background, crack, Leakage~B, Leakage~W, Leakage~G, lining falling off, and segment damage, the class weights are
\begin{equation*}
\begin{split}
\mathbf w={}&(0.000606,0.808665,0.001669,0.085032,\\
&0.05095,0.041852,0.011225).
\end{split}
\end{equation*}
They are obtained from inverse training-pixel frequencies followed by leakage-specific multipliers. The reported runs apply these weights to Dice but not to Focal.

\subsubsection{Base Segmentation Loss}\label{subsubsec:base-loss}
For each head, the class-weighted soft Dice loss is
\begin{equation}
\mathcal{L}_{\text{dice}} = \frac{1}{C}\sum_{c=1}^{C}w_c\left(1-
\frac{2\sum_i p_{i,c}g_{i,c}+\epsilon}{\sum_i p_{i,c}+\sum_i g_{i,c}+\epsilon}\right),
\label{eq:dice_loss}
\end{equation}
with $\epsilon=10^{-5}$. Classes absent from a batch remain in the class mean. The Focal term uses $\gamma=2$ and scalar $\alpha=1$:
\begin{equation}
\mathcal{L}_{\text{focal}} = -\frac{1}{N}\sum_{i=1}^{N}\sum_{c=1}^{C}
g_{i,c}\,\bigl(1-p_{i,c}\bigr)^{2}\log p_{i,c}.
\label{eq:focal_loss}
\end{equation}
The base segmentation loss combines Eqs.~\eqref{eq:dice_loss} and~\eqref{eq:focal_loss} at a $2{:}1$ ratio:
\begin{equation}
\mathcal{L}_{\text{base}} = 2.0\,\mathcal{L}_{\text{dice}} + 1.0\,\mathcal{L}_{\text{focal}}.
\label{eq:base_loss}
\end{equation}
Both terms are evaluated in float32 under mixed-precision training.

\subsubsection{Supervision of Segmentation Heads and Routing}\label{subsubsec:routing-loss}
The base loss is applied to the main fused head and to the two auxiliary morphology-specific heads introduced in Section~\ref{subsec:classification-heads}. The main head receives full supervision, while the linear and areal auxiliary heads are each weighted by $0.4$. These auxiliary terms encourage the two decoder streams to remain independently discriminative before their features are fused, preventing the branches from collapsing into redundant copies of each other.

The MRM routing map $\alpha$ (Section~\ref{subsec:mrm}) is supervised with morphologically derived binary targets at the native routing resolution ($32\times32$, matching the C3 feature scale). Ground-truth masks are downsampled by nearest-neighbour interpolation. Crack pixels form the linear set $\mathcal{S}_{L}$ with target $1$, while leakage, lining-falling-off, and segment-damage pixels form the areal set $\mathcal{S}_{A}$ with target $0$; background pixels are excluded from routing supervision. To balance the two regimes, the routing loss applies mean binary cross-entropy (mBCE) separately to the two pixel sets:
\begin{equation}
\mathcal{L}_{\text{route}} =
0.5\,\operatorname{mBCE}\!\bigl(\alpha_{\mathcal{S}_{L}},1\bigr) +
0.5\,\operatorname{mBCE}\!\bigl(\alpha_{\mathcal{S}_{A}},0\bigr),
\label{eq:routing_loss}
\end{equation}
where $\operatorname{mBCE}(\cdot,t)$ averages the binary cross-entropy between the routing weights of the indicated pixel set and the constant target $t$. The balanced set-wise construction prevents the larger morphology set from dominating the routing signal.

\subsubsection{Total Objective}\label{subsubsec:total-objective}
The complete objective aggregates the fused prediction, the two auxiliary heads, and the routing supervision:
\begin{equation}
\mathcal{L}_{\text{total}} =
\mathcal{L}_{\text{base}}^{\text{main}} +
0.4\bigl(\mathcal{L}_{\text{base}}^{\text{linear}} + \mathcal{L}_{\text{base}}^{\text{areal}}\bigr) +
0.2\,\mathcal{L}_{\text{route}}.
\label{eq:total_loss}
\end{equation}
The full model and the no-CMIM variant both use routing coefficient 0.2. The no-routing variant sets this coefficient to zero by design. Consequently, the no-CMIM comparison changes only the presence of CMIM within the controlled MDS ablation settings.

\section{Experimental Setup}\label{sec:experimental-setup}

This section defines the controlled comparison, evaluation measures, optimisation schedule, and post-training quantisation protocol used to assess MDS. All performance outcomes are reported separately in Section~\ref{sec:results}.

\subsection{Evaluation Setting and Comparators}\label{subsec:evaluation-setting}
All methods use the internal split and resolution-adaptive pipeline in Section~\ref{sec:data-pipeline}, including $512\times512$ network inputs, source-coordinate merging, metrics, and evaluation code. U-Net is randomly initialised, whereas SegDINO and MDS use DINOv3 ConvNeXt-Tiny pretraining; broad comparisons therefore combine architecture and initialisation effects. SegDINO additionally shares the encoder, augmentation, rare-class sampling, and optimisation schedule with MDS, making it the controlled comparator for morphology-aware decoding.

The representative baselines are DeepLabv3+, U-Net, SegFormer, PSPNet, and SegDINO \citep{yangSegDINOEfficientDesign2025}. SegDINO is the principal architecture-matched comparator because it uses the same DINOv3 ConvNeXt-Tiny encoder as MDS. The controlled ablation comprises the following configurations:
\begin{itemize}
\item \textbf{Baseline (SegDINO)}: DINOv3 ConvNeXt-Tiny with an MLP head, without morphology routing, dual decoders, CMIM, or routing supervision.
\item \textbf{MDS-Full}: the complete architecture with MRM, the DSA linear decoder, the PPM-FPN areal decoder, CMIM, auxiliary heads, and routing supervision.
\item \textbf{w/o CMIM}: removes CMIM while retaining routing supervision.
\item \textbf{w/o Routing}: retains the routing module but removes explicit routing supervision.
\end{itemize}

\subsection{Evaluation Metrics}\label{subsec:evaluation-metrics}
We report mean intersection over union (mIoU), mean Dice (mDice), mean per-class accuracy (mAcc), mean precision (mPrecision), and all-pixel accuracy (aAcc). Because background pixels dominate the images, mIoU is the primary comparison metric.

In addition to pixel-level metrics, we report the foreground image-level F1 ($\text{mImgF1}_{\text{fg}}$). For each foreground class $c$, an image-level true positive (TP) means that class $c$ is present in both the ground-truth and the predicted mask of an image; false positives (FP) and false negatives (FN) are defined analogously. The image-level precision, recall, and F1 are
\begin{align}
\text{ImgPrec}_c &= \frac{\text{TP}_c^{\text{img}}}{\text{TP}_c^{\text{img}} + \text{FP}_c^{\text{img}}},\label{eq:image_precision}\\
\text{ImgRec}_c &= \frac{\text{TP}_c^{\text{img}}}{\text{TP}_c^{\text{img}} + \text{FN}_c^{\text{img}}},\label{eq:image_recall}\\
\text{ImgF1}_c &= \frac{2\,\text{TP}_c^{\text{img}}}{2\,\text{TP}_c^{\text{img}} + \text{FN}_c^{\text{img}} + \text{FP}_c^{\text{img}}},\label{eq:image_f1}
\end{align}
and the foreground mean is computed over the set $C_{\text{fg}}$ of foreground classes present in the ground truth,
\begin{equation}
\text{mImgF1}_{\text{fg}} = \frac{1}{|C_{\text{fg}}|}\sum_{c\in C_{\text{fg}}}\text{ImgF1}_c.
\label{eq:mean_image_f1}
\end{equation}
This metric is relevant to engineering screening, where a missed defect category can prevent an image from being selected for human inspection.

For cracks, we further report component-level, boundary, and skeleton F1, each following
\begin{equation}
\text{F1} = \frac{2PR}{P+R},
\label{eq:diagnostic_f1}
\end{equation}
where $P$ and $R$ are computed from matched predicted and ground-truth crack components, boundary pixels, or skeleton pixels, respectively. For the leakage subclasses, the row-normalised confusion is
\begin{equation}
\text{Conf}_{i\to j} = \frac{N_{i,j}}{\sum_{j'} N_{i,j'}},
\label{eq:leakage_confusion}
\end{equation}
where $N_{i,j}$ is the number of pixels whose ground-truth leakage subclass is $i$ and whose predicted subclass is $j$.

\subsection{Implementation Details}\label{subsec:implementation-details}
MDS is implemented in PyTorch. The three training stages contain 20, 30, and 50 epochs. They respectively freeze the full backbone, retain only the earliest backbone stage frozen, and unfreeze the complete encoder. Each stage restarts a two-epoch linear warm-up followed by cosine decay of the learning rate (LR). Validation mIoU is checked every five epochs; the best checkpoints for MDS-Full, SegDINO, w/o CMIM, and w/o Routing occur at epochs 60, 95, 80, and 100, respectively.

\begin{table}[!ht]
\centering
\caption{Hyper-parameter settings for MDS.}
\label{tab:hparams}
\renewcommand{\arraystretch}{1.15}
\setlength{\tabcolsep}{6pt}
\begin{tabular}{ll}
\toprule
Hyper-parameter & Value \\
\midrule
Backbone & DINOv3 ConvNeXt-Tiny \\
Network input & $512\times512$ \\
Decoder channel & 32 \\
Optimiser & AdamW \\
Weight decay & $1\text{e-}2$ \\
Batch size & 32 \\
Base LR by stage & $10^{-3}$ / $6\!\times\!10^{-4}$ / $2\!\times\!10^{-4}$ \\
Backbone LR multiplier & 0.05 \\
LR schedule & 2-epoch warm-up + cosine \\
Stage epochs & 20 / 30 / 50 \\
Main loss & Dice + Focal \\
Auxiliary-head loss weight & 0.4 \\
Routing loss weight & 0.2 \\
Gradient clipping & 1.0 \\
\bottomrule
\end{tabular}
\end{table}

\subsection{Dynamic INT8 Quantisation Protocol}\label{subsec:quantisation-protocol}
A small on-disk footprint is a practical requirement for deploying tunnel-inspection models on resource-constrained field hardware. We therefore adopt a model-state storage budget of $\leq 50$~mebibytes (MiB) as a representative target and apply selective dynamic INT8 post-training quantisation using PyTorch with the \texttt{fbgemm} backend.

The 39 converted modules comprise 36 ConvNeXt MLP linear layers, the final DSA direction-predictor linear layer, and the first linear layer in each CMIM gate. Convolution, normalisation, activation, and attention remain floating point; no pruning or distillation is used. Section~\ref{subsec:compression-results} reports model-state size and validation mIoU before and after conversion.

\section{Results}\label{sec:results}

\subsection{Comparison with Representative Segmentation Baselines}\label{subsec:baseline-comparison}

\subsubsection{Overall Quantitative Results}\label{subsubsec:overall-results}
The same-split benchmark appears in Table~\ref{tab:overall}. All models use the same dataset split, input resolution, prediction-merging procedure, metrics, and evaluation code, while encoder initialisation follows the model-specific settings in Section~\ref{subsec:evaluation-setting}. In particular, U-Net is trained from random initialisation, whereas SegDINO and MDS use self-supervised DINOv3 ConvNeXt-Tiny pretraining. The results of the pretrained models illustrate the practical value of transferable representations on this dataset, although comparisons with the randomly initialised U-Net also contain architectural differences. SegDINO \citep{yangSegDINOEfficientDesign2025} is therefore the principal controlled comparator: it additionally shares the augmentation, rare-class sampling, optimisation schedule, and encoder used by MDS. It has 27.87~M parameters, compared with 30.00~M for MDS, enabling a controlled comparison between unified and morphology-decoupled decoding at a similar model scale.

\begin{table}[!ht]
\centering
\caption{Overall quantitative comparison under the same validation and evaluation protocol; encoder initialisation follows the model-specific settings in Section~\ref{subsec:evaluation-setting}.}
\label{tab:overall}
\renewcommand{\arraystretch}{1.15}
\setlength{\tabcolsep}{8pt}
\begin{tabular}{lccc}
\toprule
Model & mIoU (\%) & mDice (\%) & aAcc (\%) \\
\midrule
DeepLabv3+ & 54.27 & 67.48 & 84.66 \\
UNet & 23.35 & 31.82 & 76.72 \\
SegFormer & 76.20 & 86.43 & 95.87 \\
PSPNet & 71.48 & 81.67 & 94.63 \\
SegDINO & 80.08 & 87.87 & 97.87 \\
\textbf{MDS (ours)} & \textbf{81.17} & \textbf{88.47} & \textbf{98.11} \\
\bottomrule
\end{tabular}
\end{table}

The MDS run has the largest values in the table. Relative to SegDINO, the differences are $+1.09$ mIoU points, $+0.60$ mDice points, and $+0.24$ aAcc points. All-pixel accuracy is dominated by background, so the following analysis emphasises per-class IoU and foreground image-level F1.

\subsubsection{Per-class Segmentation Analysis}\label{subsubsec:perclass-results}
The first two model columns of Table~\ref{tab:ablation_perclass} compare MDS-Full with the backbone-matched SegDINO baseline at the per-class IoU level; the remaining columns support the component analysis in Section~\ref{subsubsec:perclass-ablation}.

\begin{table}[!ht]
\centering
\caption{Per-class IoU comparison of the backbone-matched baseline, full MDS, and component ablations.}
\label{tab:ablation_perclass}
\footnotesize
\renewcommand{\arraystretch}{1.2}
\setlength{\tabcolsep}{4pt}
\begin{tabular}{lcccc}
\toprule
Class & Baseline & \makecell{MDS-\\Full} & \makecell{w/o\\CMIM} & \makecell{w/o\\Routing} \\
\midrule
Background & 97.62 & \textbf{97.88} & 97.71 & 97.71 \\
Crack & 44.67 & 43.86 & \textbf{47.18} & 46.08 \\
Leakage B & 81.57 & \textbf{83.97} & 82.74 & 82.51 \\
Leakage W & 72.39 & \textbf{72.63} & 71.81 & 72.24 \\
Leakage G & 91.41 & \textbf{92.45} & 90.24 & 91.90 \\
Lining Falling Off & 80.11 & \textbf{83.59} & 80.19 & 79.72 \\
Segment Damage & 92.78 & \textbf{93.84} & 93.44 & 93.26 \\
\bottomrule
\end{tabular}
\end{table}

MDS is higher on 6 of the 7 classes. The largest gains occur for lining falling off (3.48 points), Leakage~B (2.40 points), and segment damage (1.06 points), demonstrating the benefit of morphology-aware decoding for spatially extensive defects while maintaining strong performance across the complete class set.

\subsubsection{Image-level Defect Detection}\label{subsubsec:image-level-results}
Table~\ref{tab:imagef1} reports whether each foreground class is detected at image level, irrespective of exact pixel overlap.

\begin{table}[!ht]
\centering
\caption{Per-class image-level F1 comparison for foreground defect detection.}
\label{tab:imagef1}
\renewcommand{\arraystretch}{1.15}
\setlength{\tabcolsep}{8pt}
\begin{tabular}{lcc}
\toprule
Class & SegDINO F1 (\%) & MDS F1 (\%) \\
\midrule
Crack & \textbf{88.57} & 87.67 \\
Leakage B & \textbf{97.32} & 96.71 \\
Leakage W & 53.33 & \textbf{71.43} \\
Leakage G & 50.00 & 50.00 \\
Lining Falling Off & 74.07 & \textbf{83.33} \\
Segment Damage & 87.50 & \textbf{93.02} \\
Foreground mean & 75.13 & \textbf{80.36} \\
\bottomrule
\end{tabular}
\end{table}

The foreground macro-mean is 5.23 points higher for MDS. The largest gains occur for Leakage~W (18.10 points), lining falling off (9.26 points), and segment damage (5.52 points), while Leakage~G remains comparable. A class is counted as present when at least one pixel of that class occurs in the mask; no minimum-area threshold is applied.

\subsubsection{Qualitative Comparison}\label{subsubsec:qualitative-results}
\begin{figure*}[width=.92\textwidth,pos=t]
  \centering
  \includegraphics[width=.92\linewidth]{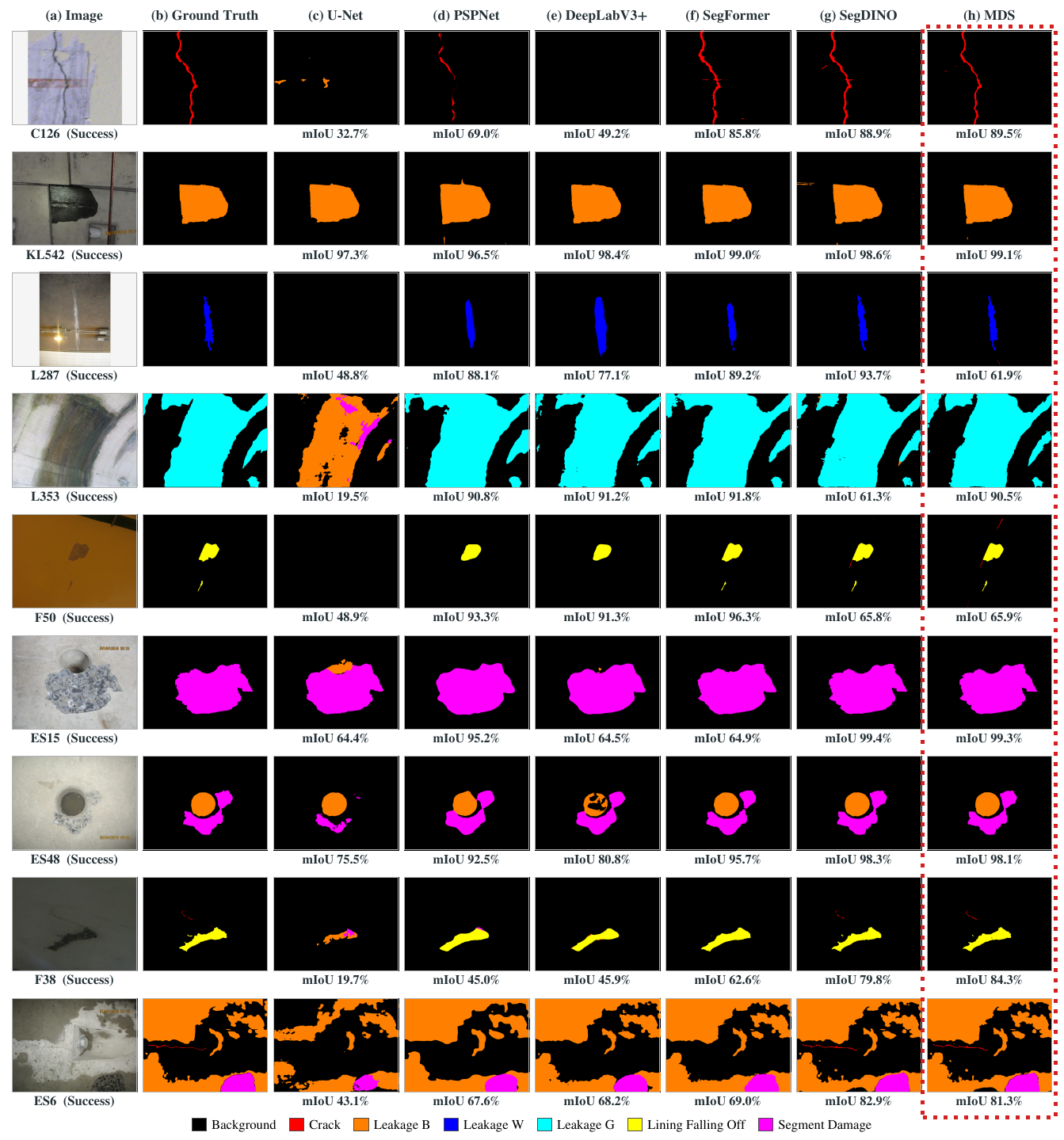}
  \caption{Nine validation cases: (a) input, (b) ground truth, and predictions from (c) U-Net, (d) PSPNet, (e) DeepLabV3+, (f) SegFormer, (g) SegDINO, and (h) MDS. Values are per-image mIoU; the red frame marks MDS.}
  \label{fig:qualitative}
\end{figure*}

Figure~\ref{fig:qualitative} shows that MDS generally produces more complete areal masks and preserves crack continuity, including in co-occurring cases. SegDINO can remain competitive on isolated leakage tiles, consistent with the crack--areal trade-off examined in Section~\ref{subsubsec:crack-diagnostics}.

\subsection{Ablation Study}\label{subsec:ablation-results}
\subsubsection{Aggregate Contribution of MDS Components}\label{subsubsec:aggregate-ablation}
Table~\ref{tab:ablation_agg} reports metrics for the four configurations.

\begin{table}[!ht]
\centering
\caption{Aggregate ablation results for the MDS components under the same evaluation protocol.}
\label{tab:ablation_agg}
\footnotesize
\renewcommand{\arraystretch}{1.2}
\setlength{\tabcolsep}{4pt}
\begin{tabular}{lcccc}
\toprule
Metric (\%) & Baseline & \makecell{MDS-\\Full} & \makecell{w/o\\CMIM} & \makecell{w/o\\Routing} \\
\midrule
mIoU & 80.08 & \textbf{81.17} & 80.47 & 80.49 \\
mIoU$_{\text{fg}}$ & 77.16 & \textbf{78.39} & 77.60 & 77.62 \\
mDice & 87.87 & \textbf{88.47} & 88.23 & 88.18 \\
mDice$_{\text{fg}}$ & 86.05 & \textbf{86.73} & 86.46 & 86.40 \\
mAcc & 86.16 & \textbf{86.64} & 86.29 & 85.56 \\
mPrecision & 89.94 & 90.79 & 90.55 & \textbf{91.30} \\
aAcc & 97.87 & \textbf{98.11} & 97.96 & 97.94 \\
mImgF1$_{\text{fg}}$ & 75.13 & \textbf{80.36} & 75.32 & 74.17 \\
\bottomrule
\end{tabular}
\end{table}

MDS-Full leads most aggregate metrics and improves $\text{mImgF1}_{\text{fg}}$ by $5.23$ points over the baseline. Removing CMIM or routing supervision reduces mIoU by $0.70$ or $0.68$ points and $\text{mImgF1}_{\text{fg}}$ by $5.04$ or $6.19$ points, respectively. The w/o Routing variant has the highest mPrecision but lower mAcc and image-level F1.

\subsubsection{Per-class Ablation Analysis}\label{subsubsec:perclass-ablation}
The full model is highest on all five areal categories. It improves lining fall-off by 3.87 points over w/o Routing and Leakage~B, Leakage~G, and lining fall-off by 1.23, 2.21, and 3.40 points over w/o CMIM, supporting the benefits of interaction and routing supervision.

\subsection{Morphology-Specific Diagnostic Analysis}\label{subsec:morphology-diagnostics}

\subsubsection{Crack Topology Metrics}\label{subsubsec:crack-diagnostics}
Crack IoU is highest for the w/o CMIM variant rather than the full model. Table~\ref{tab:crack_diag} provides a finer breakdown of crack behaviour.

\begin{table}[!ht]
\centering
\caption{Crack-specific pixel-, image-, and topology-sensitive diagnostic metrics.}
\label{tab:crack_diag}
\footnotesize
\renewcommand{\arraystretch}{1.2}
\setlength{\tabcolsep}{4pt}
\begin{tabular}{lcccc}
\toprule
Metric (\%) & Baseline & \makecell{MDS-\\Full} & \makecell{w/o\\CMIM} & \makecell{w/o\\Routing} \\
\midrule
Crack IoU & 44.67 & 43.86 & \textbf{47.18} & 46.08 \\
Crack Dice & 61.75 & 60.98 & \textbf{64.12} & 63.09 \\
Crack recall & 59.95 & 57.27 & \textbf{61.41} & 58.05 \\
Crack precision & 63.66 & 65.21 & 67.07 & \textbf{69.08} \\
Crack image F1 & 88.57 & 87.67 & \textbf{91.67} & 91.18 \\
Crack component F1 & 78.74 & 74.61 & 78.48 & \textbf{79.88} \\
Crack boundary F1 & 58.65 & 58.12 & 60.25 & \textbf{61.45} \\
Crack skeleton F1 & 65.70 & 64.70 & \textbf{66.42} & 65.42 \\
\bottomrule
\end{tabular}
\end{table}

Removing CMIM produces the highest crack IoU, Dice, recall, image F1, and skeleton F1, but reduces all leakage-subclass IoUs and lowers $\text{mImgF1}_{\text{fg}}$ from $80.36\%$ to $75.32\%$. MDS-Full instead achieves the strongest areal-class results and overall foreground image-level F1. Section~\ref{sec:discussion} examines the directional implication of this crack--areal trade-off.

\subsubsection{Leakage Subclass Confusion}\label{subsubsec:leakage-diagnostics}
The leakage subclasses are visually similar and often differ by colour, staining pattern, and environmental context. Table~\ref{tab:leakage_conf} reports leakage mIoU, leakage mDice, and the row-normalised confusion rate from Leakage~G ground truth to Leakage~B prediction. This direction is particularly diagnostic because Leakage~G is a rare subclass that can be absorbed by the dominant Leakage~B category; reporting it alongside the aggregate leakage metrics avoids duplicating the same comparison in a separate confusion-matrix figure.

\begin{table}[!ht]
\centering
\caption{Leakage-subclass metrics and Leakage~G$\to$Leakage~B confusion under different configurations.}
\label{tab:leakage_conf}
\footnotesize
\renewcommand{\arraystretch}{1.2}
\setlength{\tabcolsep}{5pt}
\begin{tabular}{lccc}
\toprule
Configuration & \makecell{Leakage\\mIoU (\%)} & \makecell{Leakage\\mDice (\%)} & \makecell{G$\to$B\\Conf.} \\
\midrule
Baseline & 81.79 & 89.78 & 0.013\% \\
MDS-Full & \textbf{83.02} & \textbf{90.50} & \textbf{0.000\%} \\
w/o CMIM & 81.60 & 89.67 & 0.149\% \\
w/o Routing & 82.21 & 90.03 & 0.745\% \\
\bottomrule
\end{tabular}
\end{table}

MDS-Full achieves the best leakage mIoU and mDice with zero Leakage~G$\to$Leakage~B confusion on the validation split, whereas removing routing supervision increases this confusion to $0.745\%$. The result associates routing supervision with more stable leakage-subclass discrimination.

\subsection{Model Size and INT8 Quantisation}\label{subsec:compression-results}
The full MDS model contains 29{,}998{,}838 trainable parameters (approximately 30.00~M), whose module-level distribution is reported in Table~\ref{tab:param_breakdown}. The parameter count implies an uncompressed 32-bit floating-point (FP32) payload of approximately 114.5~MiB.

\begin{table}[!ht]
\centering
\caption{Exact trainable-parameter breakdown of the full MDS model.}
\label{tab:param_breakdown}
\footnotesize
\renewcommand{\arraystretch}{1.15}
\setlength{\tabcolsep}{8pt}
\begin{tabular}{lr}
\toprule
Module & Parameters \\
\midrule
DINOv3 ConvNeXt-Tiny backbone & 27{,}818{,}592 \\
MRM & 1{,}634{,}689 \\
Linear/DSA decoder & 88{,}768 \\
Areal decoder & 443{,}136 \\
CMIM & 12{,}736 \\
Main classification head & 455 \\
Linear auxiliary head & 231 \\
Areal auxiliary head & 231 \\
\midrule
Total & 29{,}998{,}838 \\
\bottomrule
\end{tabular}
\end{table}

Table~\ref{tab:quantisation} shows that selective dynamic INT8 conversion reduces MDS from 114.5 to 40.5~MiB ($2.83\times$) with an mIoU change of $-0.0009$, meeting the adopted storage target. Hardware-specific latency, energy, and throughput are not measured.

\begin{table*}[pos=t]
\centering
\caption{Checkpoint size and segmentation accuracy before and after selective dynamic INT8 quantisation.}
\label{tab:quantisation}
\footnotesize
\renewcommand{\arraystretch}{1.2}
\setlength{\tabcolsep}{7pt}
\begin{tabular}{lcccccc}
\toprule
Configuration & \makecell{FP32 size\\(MiB)} & \makecell{Dynamic INT8\\size (MiB)} & \makecell{Compression\\ratio} & \makecell{FP32\\mIoU} & \makecell{Dynamic INT8\\mIoU} & $\Delta$ mIoU \\
\midrule
Baseline & 106.4 & 32.4 & $3.29\times$ & 0.8008 & 0.7984 & $-0.0024$ \\
MDS (full) & 114.5 & 40.5 & $2.83\times$ & 0.8117 & 0.8108 & $-0.0009$ \\
\bottomrule
\end{tabular}
\end{table*}

\subsection{Official Contest Evaluation}\label{subsec:contest-results}
MDS reaches 81.17\% mIoU on the internal validation split. In the 2026 Intelligent Multi-defect Detection for Tunnel Linings in Complex Environments competition, the submitted model achieved 67.55\% seven-class mIoU on the organiser-controlled hidden test, and the team placed second in the official competition ranking. Because the hidden annotations and data distribution are unavailable, the 13.62-point difference cannot be attributed to a specific cause. The hidden evaluation, organised independently by the Chinese Society of Rock Mechanics and Engineering and \emph{Underground Space}, complements the internal comparison with a sequestered test set.

\section{Discussion}\label{sec:discussion}

Morphology decoupling improves balanced multi-defect recognition rather than every class uniformly. The central question is therefore not whether an additional decoder increases aggregate capacity, but how specialisation and controlled interaction redistribute linear and areal performance.

\paragraph{Crack--areal interaction asymmetry.} MDS-Full is highest on all five areal classes and gives the strongest aggregate and image-level results, whereas removing CMIM raises crack IoU from 43.86\% to 47.18\% but lowers every areal IoU (Table~\ref{tab:ablation_perclass}). Linear features encode elongated discontinuities and boundary cues that may assist areal delineation. Conversely, broad areal texture and context may be less discriminative for the linear stream and weaken crack-specific features. A unidirectional linear-to-areal CMIM, or direction-specific gates, could preserve the former benefit without the reverse interference.

\paragraph{Efficiency-aware spatial modelling.} The MRM predicts a $32\times32$ routing map for a $512\times512$ input, DSA operates at $64\times64$, and CMIM cross-attention is pooled to $32\times32$ before gated fusion at $128\times128$. This hierarchy balances spatial detail with computational efficiency: directional modelling captures crack continuity, while pooled interaction supplies cross-morphology context at modest cost. Multi-scale routing and sparse higher-resolution interaction offer promising extensions of this design.

\paragraph{Model-state compression.} Selective dynamic INT8 conversion reduces the stored MDS state below the adopted 50~MiB budget while preserving internal-validation accuracy. This finding establishes storage compression for the converted graph, but it does not by itself establish edge-deployment performance. Hardware-specific latency, throughput, peak memory, and energy measurements remain necessary before claims about operational acceleration or efficiency can be made.

\begin{figure*}[pos=t]
  \centering
  \includegraphics[width=\linewidth]{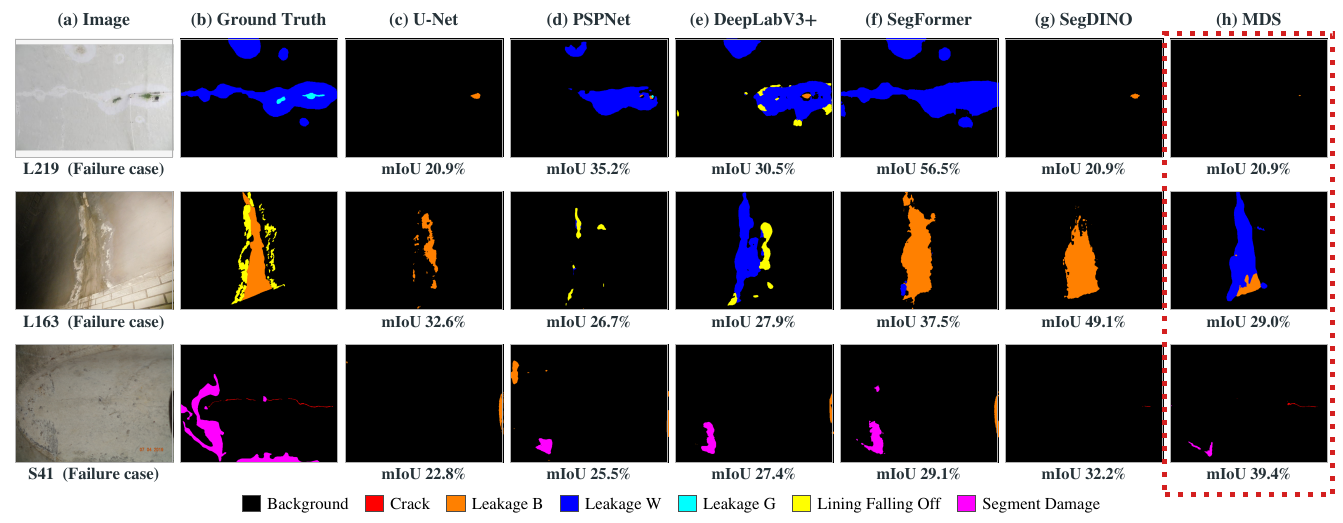}
  \caption{Failure cases: (a) input, (b) ground truth, and predictions from (c) U-Net, (d) PSPNet, (e) DeepLabV3+, (f) SegFormer, (g) SegDINO, and (h) MDS. Values are per-image mIoU; the red frame marks MDS. L219, L163, and S41 illustrate faint Leakage~W, Leakage~B/W confusion, and a crack beside segment damage.}
  \label{fig:failure}
\end{figure*}

\paragraph{Leakage subclasses.} MDS-Full achieves the highest leakage mIoU and mDice and records zero misclassification from Leakage~G into Leakage~B on the validation set. This indicates that the areal stream and routing supervision effectively preserve distinctions among visually related leakage categories. Because these categories also contain informative colour and texture cues, explicit colour-aware modelling could complement morphology routing in future extensions. Representative failure modes involving leakage appearance and co-located defects are shown in Figure~\ref{fig:failure}.

\paragraph{Limitations and future directions.} The failure cases in Figure~\ref{fig:failure} expose three complementary limitations: a faint Leakage~W region missed by all methods (L219), Leakage~B/W chromatic confusion (L163), and a thin crack lost beside a dominant segment-damage region (S41) --- respectively reflecting resolution, colour-discrimination, and crack-sensitivity constraints. Beyond these, the model has been validated on a single tunnel-inspection dataset provided by the contest organiser; evaluation on additional datasets with different lining materials, construction methods, and deterioration patterns is necessary to assess generalisability. The routing scheme currently distinguishes two morphological regimes; a finer-grained scheme with three or more regimes could better accommodate intermediate classes such as Leakage~W, whose elongation ($4.93\times$) lies between cracks and typical areal defects (Section~\ref{subsec:morphology}). Finally, post-processing methods such as crack-specific morphological filtering could refine the raw segmentation output for engineering-reporting purposes. Together, these directions point toward a more complete morphology- and appearance-aware framework for multi-defect tunnel inspection.

% Use the style of numbering in square brackets.
% If nothing is used, default style will be taken.
%\begin{enumerate}[a)]
%\item 
%\item 
%\item 
%\end{enumerate}  

%\begin{itemize}
%\item 
%\item 
%\item 
%\end{itemize}  

%\begin{description}
%\item[]
%\item[] 
%\item[] 
%\end{description}  

% Uncomment and use as the case may be
%\begin{theorem} 
%\end{theorem}

% Uncomment and use as the case may be
%\begin{lemma} 
%\end{lemma}


\section{Conclusion}\label{sec:conclusion}

This paper presents MDS, a morphology-decoupled segmentation framework for multi-defect tunnel-lining inspection under complex environments. Rather than attaching an isolated morphology-aware loss or attention mechanism to a shared segmentation pathway, MDS makes morphology the organising principle for feature routing, specialised decoding, and controlled interaction before prediction. The dataset analysis distinguishes thin linear cracks from five areal defect categories, and the model implements this separation through soft morphological routing, a DSA linear decoder, a PPM-FPN areal decoder, and gated cross-stream interaction through CMIM.

On the internal validation split, MDS reaches 81.17\% mIoU and 80.36\% foreground image-level F1, exceeding the backbone-matched SegDINO baseline by 1.09 and 5.23 percentage points, respectively. The improvement is concentrated in the areal classes, while the crack diagnostics reveal a trade-off between cross-morphology interaction and linear-stream specialisation. Selective dynamic INT8 conversion reduces model-state storage from 114.5 to 40.5~MiB while retaining 81.08\% internal-validation mIoU.

In the 2026 Intelligent Multi-defect Detection for Tunnel Linings in Complex Environments competition, the submitted model achieved 67.55\% seven-class mIoU on the organiser-controlled hidden test, and the team placed second. Together with the internal comparisons and controlled ablations, this result supports the practical competitiveness of morphology-decoupled segmentation. Validation on additional tunnel datasets and measurement on target edge hardware remain necessary to establish generalisability and deployment efficiency.

\section*{Declaration of Competing Interest}
The authors declare that they have no known competing financial interests or personal relationships that could have appeared to influence the work reported in this paper.

% Acknowledgements are omitted from the double-anonymised review manuscript.
\ifblinded\else
\section*{Acknowledgements}
We thank the competition organisers, the Chinese Society of Rock Mechanics and Engineering and \emph{Underground Space}, for the tunnel-lining data and permission to share them for research.

Funding was provided by the following programmes: Coal-Major Project, no. 2026ZD1702000; National Natural Science Foundation of China, nos. 52404242 and 52404098; China Postdoctoral Science Foundation, nos. 2024MD753977 and 2024MD753981; and Shaanxi Provincial Science Foundation, no. 2024JC-YBMS-365.
\fi

\section*{Data Availability}
The organiser-supplied images and annotations are available for research upon reasonable request from the data custodian\ifblinded\else, Mohan Ren (\texttt{mohan.ren@uq.edu.au})\fi, with acknowledgement of the competition as their source. Code and metric artifacts will be released upon acceptance and linked in the published article.

% CRediT names are hidden from the review manuscript.
\ifblinded\else\printcredits\fi

\bibliographystyle{cas-model2-names}

\bibliography{mds-refs}



\end{document}